\documentclass[runningheads]{llncs}
\usepackage{hyperref}
\hypersetup{
    colorlinks=true,
    linkcolor=blue,
    filecolor=blue,      
    urlcolor=blue,
    citecolor=blue
}
\usepackage{cite}
\usepackage{graphicx}
\usepackage{caption}
\usepackage{subfigure}
\usepackage{indentfirst} 
\usepackage{amsmath}
\usepackage{bm}
\usepackage{multirow}
\usepackage{booktabs}
\usepackage{graphicx}
\usepackage[table,xcdraw]{xcolor}
\usepackage[section]{placeins}
\usepackage{amssymb}

\usepackage[T1]{fontenc}
\usepackage{graphicx}

\makeatletter
\newcommand*\bigcdot{\mathpalette\bigcdot@{.5}}
\newcommand*\bigcdot@[2]{\mathbin{\vcenter{\hbox{\scalebox{#2}{$\m@th#1\bullet$}}}}}
\makeatother
\begin{document}
\title{Towards Generalizable Diabetic Retinopathy Grading in Unseen Domains}
\titlerunning{Generalizable Diabetic Retinopathy Grading in Unseen Domains}
%





\author{Haoxuan Che\inst{1} \and Yuhan Cheng\inst{1} \and Haibo Jin\inst{1} \and Hao Chen\thanks{Corresponding author: Hao Chen, email: jhc@cse.ust.hk.}\inst{1,2}} 
    
\authorrunning{H. Che, and et al.}

\institute{Department of Computer Science and Engineering \and Department of Chemical and Biological Engineering \\
The Hong Kong University of Science and Technology, Kowloon, Hong Kong \\
\email{\{hche, ychengbj, hjinag, jhc\}@cse.ust.hk}}

\maketitle              

\begin{abstract}

Diabetic Retinopathy (DR) is a common complication of diabetes and a leading cause of blindness worldwide.
Early and accurate grading of its severity is crucial for disease management.
Although deep learning has shown great potential for automated DR grading, its real-world deployment is still challenging due to distribution shifts among source and target domains, known as the domain generalization problem.
Existing works have mainly attributed the performance degradation to limited domain shifts caused by simple visual discrepancies, which cannot handle complex real-world scenarios. 
Instead, we present preliminary evidence suggesting the existence of three-fold generalization issues: visual and degradation style shifts, diagnostic pattern diversity, and data imbalance.
To tackle these issues, we propose a novel unified framework named Generalizable Diabetic Retinopathy Grading Network (GDRNet). 
GDRNet consists of three vital components: fundus visual-artifact augmentation (FundusAug), dynamic hybrid-supervised loss (DahLoss), and domain-class-aware re-balancing (DCR). 
FundusAug generates realistic augmented images via visual transformation and image degradation, while DahLoss jointly leverages pixel-level consistency and image-level semantics to capture the diverse diagnostic patterns and build generalizable feature representations.
Moreover, DCR mitigates the data imbalance from a domain-class view and avoids undesired over-emphasis on rare domain-class pairs.
Finally, we design a publicly available benchmark for fair evaluations.
Extensive comparison experiments against advanced methods and exhaustive ablation studies demonstrate the effectiveness and generalization ability of GDRNet.
The source code is released at \url{https://github.com/chehx/DGDR}.

\end{abstract}
\section{Introduction}
Diabetic Retinopathy (DR) is a leading cause of blindness, affecting millions of people worldwide, and early severity grading is vital for disease management \cite{cho2018idf}. 
Although deep learning (DL) has shown promising results in automatic DR grading  \cite{he2020cabnet,liu2020green, beede2020human,bai2022transformer,bai2021influence}, its real-world deployment is still challenging. 
For instance, Google's DR grading system performed ideally in controlled lab settings \cite{beede2020human}, but failed to generalize well to complex scenarios which suffer from data shifts \cite{heaven2020google}. 
It is a common problem known as domain generalization (DG) \cite{wang2022generalizing}, where the model performance significantly drops when applied to unseen domains different from the training data. 
Such an issue hinders the wide adoption and success of DL-based diagnostic tools in clinical practice \cite{li2021applications}.

Recently, several studies have explored the DG problem and reported significant performance drops in the retinal vessel segmentation \cite{wang2020dofe,zhang2020generalizing,lyu2022aadg}. 
Similarly, in DR grading, previous works showed a significant decrease in performance when presented with unseen domains and attempted to solve this problem through the perspective of feature disentanglement \cite{che2022learning} and domain-invariant feature learning \cite{atwany2022drgen}. 
Although these methods have improved performance towards unseen domains, they may not be effective in more complex real-world scenarios because they attribute the generalization issue only to limited domain shifts, such as simple visual discrepancies. 
However, the generalization issues across domains cannot be solely attributed to visual discrepancies \cite{yang2022multi}.

\begin{figure}[t]
    \centering
    \includegraphics[width=1\textwidth]{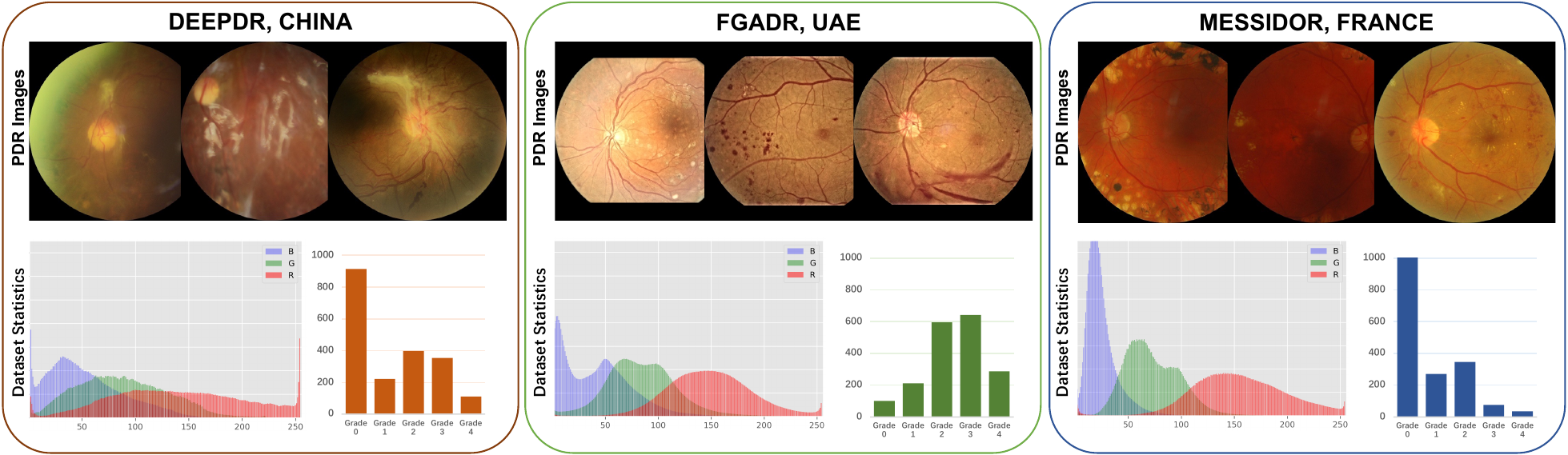}
    \caption{The RGB statistics, category histograms and proliferative DR (PDR) samples from different datasets/domains. 
    It can be observed that the existence of visual differences, image degradations and diverse diagnostic patterns from PDR samples and RGB statistics. 
    Besides, the divergence among label histograms shows the data imbalance problem across domains and categories.}
    \label{fig:domainshift}
\end{figure}

In contrast to previous works, we argue that three factors contribute to poor generalization in DGDR: visual and degradation style shifts, diagnostic pattern diversity, and data imbalance.
Specifically, as shown in Fig. \ref{fig:domainshift}, we first conduct a preliminary analysis of three public datasets/domains.
First, style shifts arise due to various factors, not only limited to visual style discrepancies, but also factors such as variations in lighting conditions \cite{shen2020modeling}, image resolution \cite{wang2021deep}, or the presence of artifacts or noise \cite{liu2022understanding}. 
These factors have been neglected in previous works yet are essential for building a generalizable model. 
Second, domains may contain diverse diagnostic patterns, such as variations in lesion types, distribution, combination and severity in certain categories \cite{liu2022deepdrid, abramoff2016improved}.
This diversity makes learning a generalizable model towards unseen domains challenging because they may contain partially-overlapped or even unknown diagnostic patterns.
Finally, data imbalance across categories and domains causes samples from specific datasets and minority classes to be underrepresented.
Moreover, this imbalance can exacerbate the issue of omitting rare diagnostic patterns, leading to shortcuts in learning and poor generalization \cite{geirhos2020shortcut}.

In this paper, we propose a novel framework, Generalizable Diabetic Retinopathy Grading Network (GDRNet) to address the DGDR problem.
Our framework consists of three critical components: fundus visual-artifact augmentation (FundusAug), dynamic hybrid-supervised loss (DahLoss), and domain-class-aware re-balancing (DCR).
By simulating visual transformations and image degradations, FundusAug enables the model to learn robust features that are less sensitive to style shifts caused by factors such as lighting conditions or artifacts and noise.
DahLoss employs a hybrid-supervised learning paradigm to handle diagnostic pattern diversity and dynamically balances the influence of supervised and unsupervised learning.
Jointly functioning with FundusAug, it enables the model to preserve pixel-level diagnostic information and learn generalizable features with sufficient intra-class variations.
Furthermore, DCR assigns soft-balanced weights to each domain-class pair to prevent underrepresentation caused by data imbalance while avoiding undesired over-emphasis introduced by hard weighting. 
Finally, to evaluate generalization ability, we design a publicly available benchmark named Generalizable Diabetic Retinopathy Grading Benchmark (GDRBench), comprising eight popular datasets and two evaluation settings. 

\begin{figure}[t]
    \centering
    \includegraphics[width=1\textwidth]{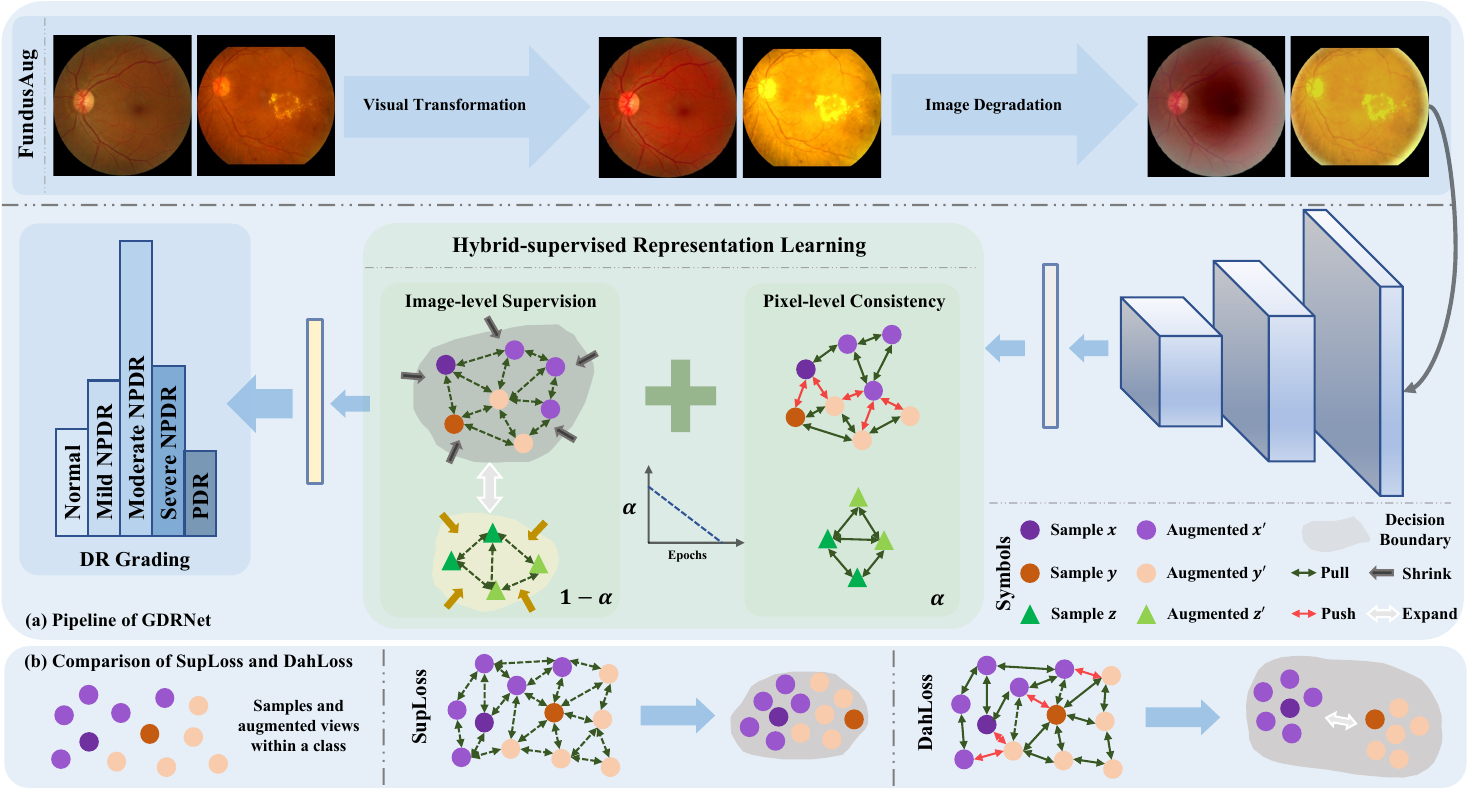}
    \caption{The pipeline of GDRNet and a high-level visual understanding of DahLoss. 
    FundusAug generates diverse, realistic augmented views, then leverages DahLoss to preserve pixel-level diagnostic patterns and learn generalizable features with sufficient intra-class variations. 
    Moreover, DCR is applied to prevent minority classes from being underrepresented. 
    }
    \label{fig:framework}
\end{figure}

\section{Methodology}

An overview of GDRNet is shown in Figure \ref{fig:framework}.
It addresses the mentioned generalization issues, including style shifts, diagnostic pattern diversity, and domain-class data imbalance, by the proposed FundusAug, DahLoss, and DCR, respectively. 
Overall, GDRNet provides a unified solution to improve the generalization ability in unseen domains. This section will introduce each component in detail.

\subsubsection{Fundus visual-artifact augmentation.}
The external machine and internal retinal illumination conditions can cause differences in visual attributes \cite{shen2020modeling}, such as contrast and brightness, while image degradations, like spot artifacts and blurring, are also very common in fundus imaging \cite{liu2022understanding,che2023image}, as depicted in Fig. \ref{fig:domainshift}.
To bridge the style shifts caused by visual discrepancies and image degradations, we developed FundusAug, which is parameter-free and plug-and-use, designed to generate diverse and realistic augmented views.
It employs five basic image transformations, including brightness, contrast, saturation, hue, and sharpness adjustments, to fill visual discrepancy gaps. 
Moreover, it uses extra four degradation-based image transformations, including halo simulation, hole generation, spot addition, and image blur, to address image degradation gaps. 
Specifically, given training data $x$ with label $y$, the augmented view $\hat{x}$ of original data can be derived through the following equation:
\begin{equation}
    \hat{x} := \text{FundusAug}(x) = \pi^n_{p_n, m_n}(\pi^{n-1}_{p_{n-1}, m_{n-1}}(...\pi^1_{p_1, m_1}(x))),
\end{equation}
where $\pi^n_{p_n, m_n}$ denotes transformation $n$ with probability $p_n$ and random  intensity $m_n$.
To reduce the parameter space of FundusAug while still ensuring image diversity, we implemented FundusAug by applying each operation with a parameter-free procedure that uniformly selects a process with a probability of 0.5. 
FundusAug can generate realistic augmented views while preserving their diagnostic semantics, as well as providing a robust foundation for subsequent generalizable feature learning by increasing image diversity.
A detailed description and visualization of operations can be found in the appendix.

\subsubsection{Dynamic hybrid-supervised loss.}
While the supervised loss (SupLoss), e.g., the cross-entropy loss (CE), effectively guides the model to learn effective feature representations \cite{zhang2021unleashing}, it has two disadvantages in DGDR. 
First, SupLoss leads to dense features within categories while sufficient intra-class variations are crucial for effectively generalizing to unseen domains \cite{duboudin2021encouraging}. 
Second, the potential variety of diagnostic patterns in unseen domains requires the models to learn pixel-level lesion semantics as much as possible, while SupLoss lacks such functionality \cite{islam2021broad}.
To tackle these issues, we proposed DahLoss to encourage models to learn features with sufficient intra-class variations and preserve diagnostic patterns, by introducing a hybrid-supervised paradigm to jointly leverage image-level severity supervision and pixel-level semantics consistency.
A basic form of DahLoss is as
\begin{equation}
    \mathcal{L}_{dhl} = (1-\alpha) \mathcal{L}_{sup} + \alpha \mathcal{L}_{scon},
\end{equation}
where $\alpha$ decreasing within range $[0, 1]$ is a hyper-parameter to dynamically control the task focus, and $\mathcal{L}_{sup}$ and $\mathcal{L}_{scon}$ could be any supervised and self-supervised contrastive loss functions.
In this paper, we adopt CE and instance discrimination loss \cite{chen2020simple} as $\mathcal{L}_{sup}$ and $\mathcal{L}_{scon}$. 
Specifically, within a multiviewed batch, let $i \in I \equiv \{1...N\}$ be the index of arbitrary augmented samples by FundusAug and $j(i) \in J \equiv \{1^\prime...N^\prime\} $ be the index of weakly-augmented samples originating from the same source sample, we denote $\mathcal{L}_{dhs}^i$ for sample $i$ as
\begin{equation}
    \mathcal{L}_{dhl}^i = - (1-\alpha) \log p_t^i - \alpha \log \frac{\exp(f_i \bigcdot f_{j(i)} / \tau)}{\sum_{a\in A(i)} \exp(f_i \bigcdot f_a) / \tau)},
\end{equation}
where $p_t^{\cdot}$ denotes the predicted probability of the true class under one-hot encoding label, $f_\cdot$ denotes the $l_2$ normalized feature, the $\bigcdot$ symbol denotes the inner product, $\tau$ is the temperature parameter and $A(i) \equiv (J \cup I) / {i}$.

As illustrated in Fig. \ref{fig:framework} (b), the use of $\mathcal{L}_{sup}$ alone tends to force samples within the same class to cluster tightly together, resulting in a highly concentrated feature representation that may not be beneficial for generalization to unseen domains.
However, by incorporating $\mathcal{L}_{scon}$, DahLoss can achieve a balance between the intra-class variation and inter-class distance in learned feature representations. 
Specifically, while $\mathcal{L}_{sup}$ encourages samples within the same class to cluster together, $\mathcal{L}_{scon}$ simultaneously pulls the augmented views of these samples closer to each other and pushes them far away from those of other samples.
It results in a feature representation with both sufficient intra-class variation and clear inter-class separation.
Moreover, DahLoss also leverages pixel-level consistency to preserve crucial diagnostic patterns. 
It is achieved by enforcing the model to maintain feature representations with semantics similarity between strong-weak augmented views originating from the same sample, which ensures that critical details and structures are learned. 
By jointly considering both image-level supervision and pixel-level consistency, DahLoss encourages the model to learn features with sufficient intra-class variations and preserve crucial diagnostic information, improving generalization performance in unseen domains.
Finally, gradually decreasing the value of $\alpha$ during training would guide the model to focus more on the grading task, leading to a balance between representation learning and grading performance. 
In this paper, we simply set $\alpha$ decay from 1 to 0 linearly across training epochs to verify the effectiveness of DahLoss.

\subsubsection{Domain-class-aware re-balancing.}
The domain-class data imbalance can result in certain categories and diagnostic patterns in specific datasets being underrepresented, leading to biased and inaccurate model predictions \cite{yang2022multi}. 
To complicate the situation, adopting a hard balancing style, such as weighting domain-class pairs based on the reciprocal of occurrence frequency, could potentially cause performance degradation \cite{ren2018learning}. 
Such decay is owing to the fact that underrepresented domain-class pairs often have a small ratio, resulting in excessively high weights to dominate gradients. 
Inspired by \cite{conneau2019cross}, we designed the DCR method to address this issue.
It assigns weights to each sample based on the occurrence probability $q_c^d$ of its category $c$ and domain $d$. 
Specifically, DCR calculates the weight $w_c^{d}$ for category $c$ in domain $d$ as
\begin{equation}
w_c^{d} = \frac{\sum_{d^{\prime} \in D}\sum_{j=1}^N (q_j^{d^{\prime}})^\beta} {(q_c^{d})^\beta},
\end{equation}
where $D$ denotes the set of domains, $N$ is the amount of classes, and $\beta$ with a range of $[0,1]$ is a hyperparameter that adjusts the balancing intensity.
When $\beta$ approaches to $0$, DCR assigns weight more equally, and when $\beta$ closes to $1$, it acts more like the naive hard balancing method. 
By introducing $\beta$, DCR enables more nuanced weighting of samples based on their domain and class, reducing the risk of over-emphasizing underrepresented samples in the loss function. 
By considering the occurrence probabilities of all categories across all domains, DCR  mitigates the data imbalance problem of underrepresented class-domain pairs. 
These two make DCR an effective solution for handling domain-class data imbalance and improving the generalizability of DR grading models. 

\begin{table*}[!tbp]
        \renewcommand\arraystretch{1.2}
        \centering
        \caption{Comparison with state-of-the-art approaches under the DG test.}    
        \label{tab:comparisonsfundus}
        \resizebox{1.0\textwidth}{!}{%
        \scalebox{0.69}{
        \begin{tabular}{c|ccc|ccc|ccc|ccc|ccc|ccc|ccc}
            \toprule
             Target &\multicolumn{3}{c|}{APTOS} &\multicolumn{3}{c|}{DeepDR}  &\multicolumn{3}{c|}{FGADR}&\multicolumn{3}{c|}{IDRID} &\multicolumn{3}{c|}{Messidor}  &\multicolumn{3}{c|}{RLDR} &\multicolumn{3}{c}{Average}\\
            \hline
            Metrics & AUC &ACC & F1 & AUC &ACC & F1&  AUC &ACC & F1 & AUC &ACC & F1 & AUC &ACC & F1&  AUC &ACC & F1 &  AUC &ACC & F1\\
            \hline
            \hline
            ERM &75.0 &44.4 &38.9 &77.0 &39.5 &34.3 &66.2 &32.0 &27.1 &82.3 &\underline{50.0} &\textbf{44.1} &\underline{79.1} &60.7 &\underline{43.4} &75.9 &36.5 &35.7 &75.9 &43.8 &37.3\\
            DRGen\cite{atwany2022drgen} &\underline{79.9} &58.1 &40.2 &\underline{83.0} &38.7 &34.1 &69.4 &41.7 &24.7 &\textbf{84.7} &44.6 &37.4 &79.0 &60.1 &40.5 &\underline{79.5} &43.1 &37.0 &\underline{79.3} &47.7 &\underline{37.3}\\
            Mixup\cite{zhangmixup} &75.3 &62.6 &43.2 &75.3 &29.0 &25.2 &66.7 &42.3 &32.3 &78.8 &39.0 &27.6 &76.7 &54.7 &32.6 &76.9 &\underline{43.6} &\underline{37.7} &75.0 &45.2 &33.1\\
            MixStyle\cite{zhoudomain}  &79.0 &65.8 &39.9 &76.9 &32.9 &27.9 &71.2 &35.8 &22.7 &83.0 &\textbf{51.4} &\underline{39.2} &75.2 &62.2 &36.5 &75.5 &41.1 &31.4 &76.8 &48.2 &32.9\\
            GREEN\cite{liu2020green}  &75.1 &53.8 &38.9 &76.4 &28.1 &24.9 &69.5 &41.3 &31.5 &79.9 &41.3 &32.2 &75.8 &52.0 &36.8 &74.8 &34.0 &34.4 &75.3 &41.8 &33.1\\
            CABNet\cite{he2020cabnet} &75.8 &55.5 &39.4 &75.2 &42.7 &31.8 &73.2 &43.7 &\underline{34.8} &79.2 &44.8 &37.3 &74.2 &56.1 &34.1 &75.8 &37.0 &35.6 &75.6 &46.6 &35.5\\
            DDAIG\cite{zhou2020deep}  &78.0 &\textbf{67.1} &41.0 &75.6 &37.6 &32.2 &73.6 &42.0 &33.8 &82.1 &37.4 &27.0 &76.6 &58.4 &35.3 &75.6 &36.1 &27.7 &76.9 &46.4 &32.8\\
            ATS\cite{yang2021adversarial} &77.1 &56.9 &38.3 &79.4 &36.1 &31.6 &\underline{74.7} &\textbf{46.7} &33.4 &83.0 &41.5 &34.9 &77.2 &\underline{64.7} &35.8 &76.5 &37.4 &34.9 &78.0 &47.2 &34.8\\
            Fishr\cite{rame2022fishr}  &79.2 &66.6 &\underline{43.4} &81.1 &\underline{48.1} &34.4 &73.3 &44.4 &34.4 &82.7 &40.3 &27.6 &76.4 &\textbf{65.1} &41.1 &77.4 &36.8 &34.7 &78.4 &\underline{50.2} &35.9\\
            MDLT\cite{yang2022multi} &77.3 &57.2 &41.5 &80.0 &39.5 &\underline{36.2} &74.1 &\underline{45.7} &29.0 &81.5 &44.2 &35.4 &75.4 &58.9 &36.9 &75.7 &37.6 &35.0 &77.3 &47.2 &35.7\\
            \midrule
            \textbf{GDRNet} &\textbf{79.9} &\underline{66.8} &\textbf{46.0} &\textbf{84.7} &\textbf{53.1} &\textbf{45.3} &\textbf{80.8} &{45.3} &\textbf{39.4} &\underline{84.0} &40.3 &35.9 &\textbf{83.2} &{63.4} &\textbf{50.9} &\textbf{82.9} &\textbf{45.8} &\textbf{43.5} &\textbf{82.6} &\textbf{52.5} &\textbf{43.5}\\
            \bottomrule
            \end{tabular}
    }}
\label{tab:DG}
\end{table*}

\section{Experiments}
\subsubsection{Experimental settings, implementation details and evaluation metrics.}
To comprehensively analyze and evaluate our framework, we designed the GDRBench involving two generalization ability evaluation settings and eight popular public datasets.
First, GDRBench preserves the classic leave-one-domain-out protocol (DG test), which requires leaving one domain for evaluation and training models on the rest.
It involves six datasets, including DeepDR\cite{liu2022deepdrid}, Messidor\cite{abramoff2016improved}, IDRID\cite{porwal2018indian}, APTOS\cite{APTOS}, FGADR\cite{FGADR}, and RLDR\cite{icpr2020-LesionNet}.
Further, we designed an extreme single-domain generalization setting (ESDG test), which follows the train-on-single-domain protocol with datasets mentioned above but adds two extra large-scale datasets, DDR\cite{LI2019} and EyePACS \cite{EYEPACS} for evaluation.
It simulates real-world generalization issues, in which models are trained only on thousands of samples but are required to generalize well on one hundred thousand images from multiple hospitals and areas.
We used ResNet50 pre-trained on ImageNet as the backbone and a fully connected layer as the linear classifier. 
For evaluation, we report three critical metrics, namely accuracy (ACC), the area under the ROC curve (AUC), and macro F1-score (F1). 
We used \textbf{bold} and \underline{underline} to indicate the first- and second-highest scores.
A detailed illustration of the datasets and implementation settings can be found in the appendix.

\subsubsection{Comparison with other methods.}
We conducted a comprehensive experiment to evaluate our framework, comparing it with a vanilla baseline (ERM) and other state-of-the-art methods from various categories, including ophthalmic disease diagnosis (OSD) \cite{liu2020green, he2020cabnet}, domain generalization techniques (DGT) \cite{zhangmixup, zhoudomain, zhou2020deep, yang2021adversarial, atwany2022drgen}, and feature representation learning (FRL) \cite{yang2022multi, rame2022fishr}.
These chosen methods can be adopted in DGDR with minimal or no modifications, and their brief descriptions are in the appendix. 
We employed a standard DG augmentation pipeline \cite{yang2022multi} for all methods except DRGen with a default augmentation strategy \cite{atwany2022drgen}.
Table \ref{tab:DG} shows the quantitative results of the DG test, where the row of ``target'' indicates the test domain.
GDRNet performs better than other methods and significantly improves at least two of AUC, ACC, and F1 on all sub-tests except the test on IDRiD, whose small scale leads to unobvious diagnostic pattern diversity.
Typically, ODS methods do not consider domain shifts, and they thus fail to generalize as well as other methods.
As expected, DGT and FRL methods consistently improve the performance compared with ERM due to specific designs towards limited domain shifts. 
However, GDRNet still outperforms these methods markedly due to it handles three-fold generalization issues via increasing training diversity via realistic augmentation, learning generalizable features with sufficient intra-class variations, preserving diagnostic patterns, and rebalancing the minority of samples.
Overall, the result shows the effectiveness of GDRNet.

\begin{table*}[!tbp]
  \renewcommand\arraystretch{1}
  \centering
  \caption{Ablation studies on proposed components under the DG test.}
  \resizebox{1\textwidth}{!}{
    \begin{tabular}{l|cccc|cccccc|c}
    \toprule
    Method & VT & ID &DCR &$\mathcal{L}_{dhl}$  & APTOS & DeepDR & FGADR & IDRiD & Messidor & RLDR  & Average \\
    \midrule
    ERM & \textbf{-} & \textbf{-} & \textbf{-} & \textbf{-} &75.04 &77.02 &66.19 &82.31 &79.10 &75.86 &75.92 \\
    \midrule
    Model A & \checkmark & \textbf{-} & \textbf{-} &\textbf{-} &77.16 &80.22 &74.87 &82.83 &80.48 &80.23 &79.30 \\
    Model B & \textbf{-} & \checkmark & \textbf{-} &\textbf{-} &75.38 &79.13 &68.35 &82.72 &77.94 &78.93 &77.08 \\
    Model C & \textbf{-} & \textbf{-} &\checkmark & \textbf{-} &75.58 &78.76 &67.34 &82.59 &78.86 &78.39 &76.92 \\
    Model D & \textbf{-} & \textbf{-} &\textbf{-} &\checkmark &77.67 &81.22 &75.34 &81.98 &79.09 &80.57 &79.31 \\
    \midrule
    Model E &\checkmark & \checkmark & \textbf{-} &\textbf{-} &77.28 &83.41 &77.44 &83.63 &80.22 &80.82 &80.46 \\
    Model F &\checkmark &\checkmark &\checkmark &\textbf{-}  &77.37 &83.90
    &77.91 &83.44 &82.43 &83.33 &81.39 \\
    Model G & \checkmark & \checkmark &\textbf{-} &\checkmark &80.40 &85.07 &78.65 &83.68 &84.91 &81.97 &82.45 \\
    \midrule
    GDRNet &\checkmark &\checkmark &\checkmark  &\checkmark &77.67 &84.69 &80.78 &84.01 &83.16 &82.91 &82.57 \\
    \bottomrule
    \end{tabular}}
  \label{ablation_component}
\end{table*}

\subsubsection{Ablation study of proposed components.}
To evaluate the effectiveness of our proposed components, we conducted an extensive ablation study under the DG test and presented the AUC score achieved by different models in Table \ref{ablation_component}. 
We first examined the individual effects of FundusAug with only visual transformation (VT), FundusAug with only image degradation (ID), DCR and DahLoss (Models A-D, respectively). 
Our results demonstrate that all proposed components effectively improve model generalization performance and address the three-fold generalization issues discussed in this paper. 
We then combined these components to analyze their joint effects (Models E-G). 
Notably, DahLoss and FundusAug contribute significantly to the improvement, demonstrating their capability to increase the training diversity, handle diagnostic pattern diversity, and dynamically balance the influence of supervised and unsupervised learning. 
Finally, we achieved the best performance by combining all the components in GDRNet, showing that our components play complementary roles.
It is worth mentioning that we can observe different performance trends across the various datasets in the ablation study, which indicates the complexity of generalization issues in DGDR. 
It also suggests that different domains may have varying grades of severity in the three-fold generalization issues identified in this paper.

\subsubsection{Generalization from a single source domain.}
Further, to comprehensively investigate the generalization performance, we introduced the ESDG test.
It is a more demanding evaluation setting than the DG test because it requires models trained on a single source dataset to generalize to new domains with significantly larger scales and different data distributions.
In contrast, the DG test involves training on multiple domains to fill domain gaps naturally.
The quantitative results are presented in Table \ref{tab:SDG}, where the row of ``source'' indicates the training dataset. 
As expected, the average performance of all methods decreases significantly, indicating the difficulty of ESDG.
Although some methods outperform others in the DG test, such as MixStyle, due to their specific design to leverage the discrepancy of domains, they fail to generalize in the strict ESDG test.
Despite more strict requirements, GDRNet still outperforms other methods in average performance and at least one metric in all sub-tests, owing to its effective designs towards three-fold generalization issues. 

\begin{table*}[!tbp]
        \renewcommand\arraystretch{1.2}
        \centering
        \caption{Comparison with state-of-the-art approaches under the ESDG test.}
        \label{tab:comparisonsfundus}
        \resizebox{1.0\textwidth}{!}{%
        \scalebox{0.69}{
        \begin{tabular}{c|ccc|ccc|ccc|ccc|ccc|ccc|ccc}
            \toprule
             Source &\multicolumn{3}{c|}{APTOS} &\multicolumn{3}{c|}{DeepDR}  &\multicolumn{3}{c|}{FGADR}&\multicolumn{3}{c|}{IDRID} &\multicolumn{3}{c|}{Messidor}  &\multicolumn{3}{c|}{RLDR} &\multicolumn{3}{c}{Average}\\
            \hline
            Metrics & AUC &ACC & F1 & AUC &ACC & F1&  AUC &ACC & F1 & AUC &ACC & F1 & AUC &ACC & F1&  AUC &ACC & F1 &  AUC &ACC & F1\\
            \hline
            \hline
            ERM &66.4 &53.2 &31.6&70.7 &47.3 &31.2 &55.3 &5.6 &7.1 &69.6 &56.5 &\underline{33.9} &70.6 &51.3 &33.7 &70.1 &27.3 &26.4 &67.1 &40.2 &27.3\\
            DRGen\cite{atwany2022drgen} &\underline{69.4} &\underline{60.7} &\textbf{35.7} &\textbf{78.5} &39.4 &31.6 &59.8 &6.8 &\underline{8.4} &70.8 &\underline{67.7} &30.6 &\underline{77.0} &64.5 &\underline{37.4} &\underline{78.9} &19.0 &21.2 &\underline{72.4} &43.0 &27.5\\
            Mixup\cite{zhangmixup} &65.5 &49.4 &30.2 &70.7 &49.7 &33.3 &58.8 &5.8 &7.4 &70.2 &64.0 &32.6 &71.5 &63.0 &32.6 &72.9 &27.7 &27.0 &68.3 &43.3 &27.2\\
            MixStyle\cite{zhoudomain} &62.0 &48.8 &25.0 &53.3 &32.0 &14.6 &51.0 &{7.0} &7.9 &53.0 &53.5 &19.4 &51.4 &57.6 &16.8 &53.5 &18.3 &6.4 &54.0 &36.2 &15.0\\
            GREEN\cite{liu2020green} &67.5 &52.6 &33.3 &71.2 &44.6 &31.1 &58.1 &5.7 &6.9 &68.5 &60.7 &33.0 &71.3 &54.5 &33.1 &71.0 &\underline{31.9} &{27.8} &67.9 &41.7 &{27.5}\\
            CABNet\cite{he2020cabnet} &67.3 &52.2 &30.8 &70.0 &\underline{55.4} &32.0 &57.1 &6.1 &7.5 &67.4 &62.7 &31.7 &72.3 &63.8 &35.3 &75.2 &23.0 &25.4 &68.2 &43.8 &27.2\\
            DDAIG\cite{zhou2020deep} &67.4 &48.7 &31.6 &73.2 &38.5 &29.7 &59.9 &5.0 &5.5 &70.2 &60.2 &33.4 &73.5 &\textbf{69.1} &35.6 &74.4 &25.4 &23.5 &69.8 &41.2 &26.7\\
            ATS\cite{yang2021adversarial} &68.8 &51.7 &32.4 &72.7 &52.4 &{33.5} &\underline{60.3} &5.3 &5.7 &69.1 &66.6 &30.6 &73.4 &64.8 &32.4 &75.0 &24.2 &23.9 &69.9 &\underline{44.2} &26.4\\
            Fishr\cite{rame2022fishr} &64.5 &\textbf{61.7} &31.0 &72.1 &\textbf{61.0} &30.1 &56.3 &6.0 &7.2 &{71.8} &48.0 &30.6 &74.3 &52.0 &33.8 &78.6 &19.3 &21.3 &69.6 &41.3 &25.7\\
            MDLT\cite{yang2022multi} &67.6 &53.3 &32.4 &73.1 &50.2 &\underline{33.7} &57.1 &\underline{7.1} &7.8 &\underline{71.9} &61.7 &32.4 &73.4 &58.9 &34.1 &76.6 &29.0 &\underline{30.0} &70.0 &43.4 &\underline{28.4}\\
            \midrule
            \textbf{GDRNet} &\textbf{69.8} &52.8 &\underline{35.2} &\underline{76.1} &40.0 &\textbf{35.0} &\textbf{63.7} &\textbf{7.5} &\textbf{9.2} &\textbf{72.9} &\textbf{70.0} &\textbf{35.1} &\textbf{78.1} &\underline{65.7} &\textbf{40.5} &\textbf{79.7} &\textbf{44.3} &\textbf{37.9} &\textbf{73.4} &\textbf{46.7} &\textbf{32.2}\\
            \bottomrule
            \end{tabular}
    }}
\label{tab:SDG}
\end{table*}

\section{Conclusion}
In this paper, we tackled the three-fold generalization issues that hinder the generalizability of DR grading, including style shifts, diagnostic pattern diversity, and data imbalance. 
To overcome these challenges, we proposed a novel and unified framework called GDRNet, incorporating three effective components: FundusAug, DahLoss, and DCR.
FundusAug enables the generation of realistic augmented views, while DahLoss leverages supervised and unsupervised learning to preserve diagnostic patterns and increase intra-class variations of features.
Finally, DCR softly handles the data imbalance across categories and domains to avoid potential performance decay.
Together, these three components work synergistically to improve the generalization performance of the model.
GDRNet achieved superior performance on both DG and ESDG tests of the proposed publicly available GDRBench, demonstrating its effectiveness and robustness in addressing the three-fold generalization issues in DR grading.
Overall, our work provides valuable insights and practical solutions for improving the generalization capability of deep learning in medical image analysis, and has the potential to benefit real-world clinical applications.

\textbf{Acknowledgement.} 
This work was supported by the Hong Kong Innovation and Technology Fund (Project No. ITS/028/21FP), Shenzhen Science and Technology Innovation Committee Fund (Project No. SGDX20210823103201011) and HKUST 30 for 30 Research Initiative Scheme.

%
%
\bibliographystyle{splncs04}
\bibliography{paper1591}


\end{document}


\title{Appendix for ``Towards Generalizable Diabetic Retinopathy Grading in Unseen Domains''}  

\titlerunning{Generalizable Diabetic Retinopathy Grading in Unseen Domains}

\author{Haoxuan Che \and Yuhan Cheng \and Haibo Jin \and Hao Chen} 

\authorrunning{H. Che, et al.}

\institute{The Hong Kong University of Science and Technology}




\maketitle              


\begin{figure}
    \vspace{-4mm}
    \centering
    \includegraphics[width=1\textwidth]{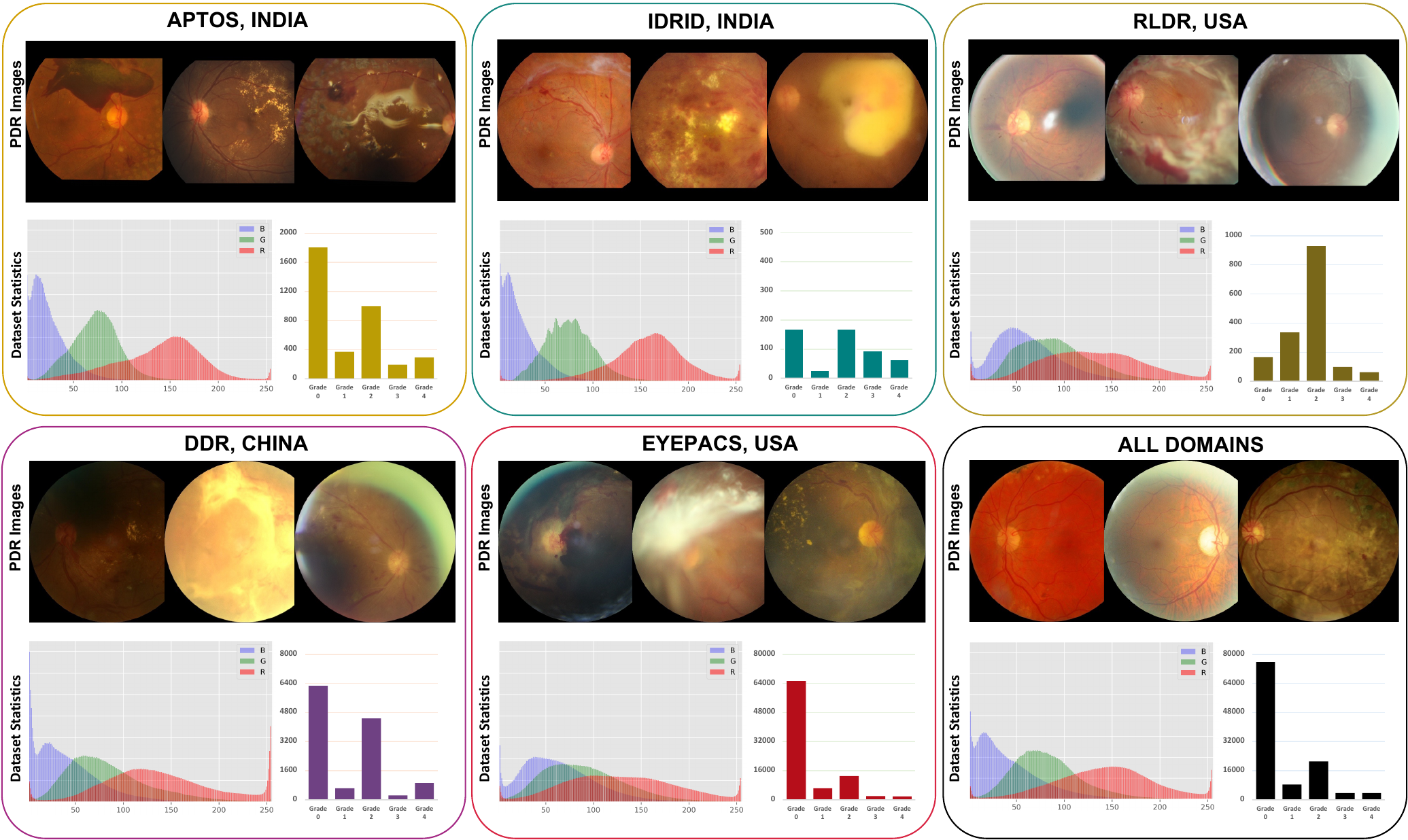}
    \caption{\small{The RGB statistics, category histograms and proliferative DR (PDR) samples from other five datasets/domains.}}
    \label{fig:supp_domain}
    \vspace{-4mm}
\end{figure}

\begin{figure}
    \vspace{-4mm}
    \centering
    \includegraphics[width=1.0\textwidth]{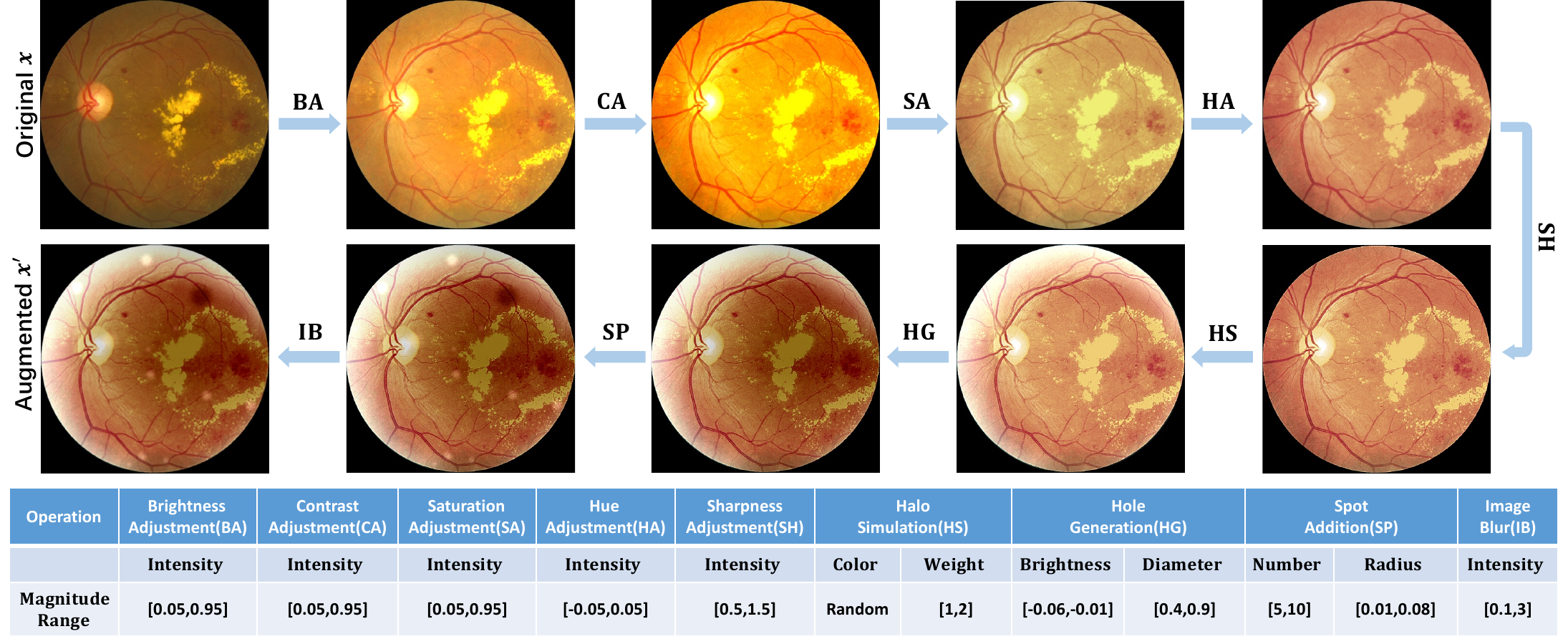}
    \caption{\small{Augmentation pipelines and parameter ranges of all FundusAug.}}
    \label{fig:operations}
    \vspace{-4mm}
\end{figure}

\begin{figure}
    \centering
    \includegraphics[width=1\textwidth]{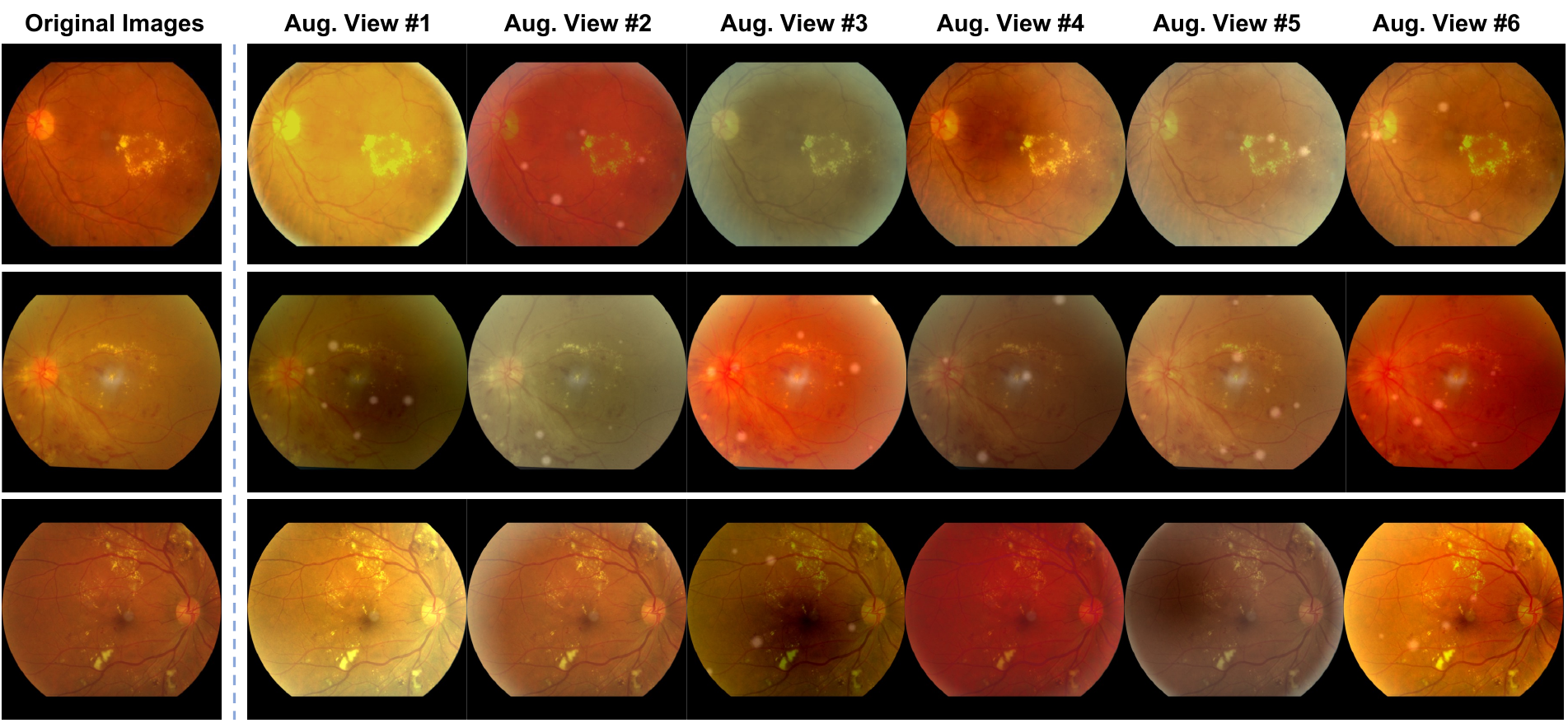}
    \caption{\small{Augmented samples via FundusAug. These views show the effectiveness of FundusAug in increasing the training sample diversity.}}
    \label{fig:augmentedviews}
    \vspace{-4mm}
\end{figure}

\begin{table*}
    \renewcommand\arraystretch{1.1}
    \centering
    \caption{\small{Controllable parameters during training.}}
    \label{tab:fundusparameters}   
    \resizebox{1.0\textwidth}{!}{
    \begin{tabular}{c|c|c|c|c|c|c|c}
    \toprule
    Parameters & Epochs &Learning rate (decayed) & Batch size & Weight decay & Momentum value & Intensity $\beta$ & Temperature $\tau$ \\
    \hline
    \hline
    Comparison methods &100 &$10^{-3}$ ($10^{-4}$) &$16$ &$5\times10^{-4}$ &$0.9$ &\textbf{-} &\textbf{-} \\
    GDRNet (DG)  &100 &$10^{-3}$ ($10^{-4}$) &$16$ &$5\times10^{-4}$ &$0.9$ &$0.5$ &$0.1$ \\
    GDRNet (ESDG) &100 &$10^{-3}$ ($10^{-4}$) &$16$ &$5\times10^{-4}$ &$0.9$ &$0.5$ &$0.01$ \\
  
    \bottomrule
    \end{tabular}
    }
    \vspace{-4mm}
\end{table*}

\begin{table*}
    \renewcommand\arraystretch{1.1}
    \centering
    \caption{\small{A brief summary for comparison methods.}}
    \label{tab:comparisonsfundus}   
    \resizebox{1.0\textwidth}{!}{%
    \begin{tabular}{c|c|l}
    \toprule
    \multicolumn{1}{c|}{Name} &  \multicolumn{1}{c|}{Category} & \multicolumn{1}{c}{Brief Description} \\
        
    \hline
    GREEN & OSD & \multicolumn{1}{m{12cm}}{Utilize class dependency priors for DR grading via a graph convolutional network.} \\
    \hline
    CABNet & OSD &  \multicolumn{1}{m{12cm}}{Learn discriminative features for each grade via the categorical attention block.} \\
    \hline
    Mixup & DGT &  \multicolumn{1}{m{12cm}}{Mix both images and labels to improve generalization ability.} \\
    \hline
    MixStyle & DGT &  \multicolumn{1}{m{12cm}}{Mix feature statistics of training samples across domains to improve generalization.} \\    
    \hline
    DDAIG & DGT &  \multicolumn{1}{m{12cm}}{Generate novel perturbed images to improve generalization via adversarial training.} \\    
    \hline
    ATS & DGT &  \multicolumn{1}{m{12cm}}{Advances Teacher-Student learning in an adversarial learning manner.} \\    
    \hline
    DRGen & DGT &  \multicolumn{1}{m{12cm}}{Combine Fishr and SWAD, which aims to find the flat optima.} \\    
    \hline
    Fishr & FRL &  \multicolumn{1}{m{12cm}}{Enforce domain invariance in the space of the gradients of the loss.} \\    
    \hline
    MDLT & FRL &  \multicolumn{1}{m{12cm}}{Based on the prototype learning, address data imbalance and domain shifts.} \\    
    \bottomrule
    \end{tabular}
    }
    \vspace{-4mm}
\end{table*}

\begin{table*}
        \renewcommand\arraystretch{1.2}
        \centering
        \caption{\small{Ablation studies on proposed components under the DG test.}}
        \label{tab:comparisonsfundus}
        \resizebox{1.0\textwidth}{!}{%
        \scalebox{0.69}{
        \begin{tabular}{c|ccc|ccc|ccc|ccc|ccc|ccc|ccc}
        \toprule
             Target &\multicolumn{3}{c|}{APTOS} &\multicolumn{3}{c|}{DeepDR}  &\multicolumn{3}{c|}{FGADR}&\multicolumn{3}{c|}{IDRID} &\multicolumn{3}{c|}{Messidor}  &\multicolumn{3}{c|}{RLDR} &\multicolumn{3}{c}{Average}\\
            \hline
            Metrics & AUC &ACC & F1 & AUC &ACC & F1&  AUC &ACC & F1 & AUC &ACC & F1 & AUC &ACC & F1&  AUC &ACC & F1 &  AUC &ACC & F1\\
            \hline
            \hline
            ERM &75.0 &44.4 &38.9 &77.0 &39.5 &34.3 &66.2 &32.0 &27.1 &82.3 &50.0 &44.1 &79.1 &60.7 &43.5 &75.9 &36.5 &35.7 &75.9 &43.8 &37.3\\
            \midrule
            Model A &77.2 &53.8 &40.6 &80.2 &42.0 &37.7 &74.9 &42.7 &37.7 &82.8 &53.1 &45.3 &80.5 &61.1 &47.3 &80.2 &43.6 &40.0 &79.3 &49.4 &41.4\\
            Model B &75.4 &38.6 &32.7 &79.1 &41.4 &34.3 &68.4 &33.0 &27.5 &82.7 &50.2 &45.1 &77.9 &62.6 &45.8 &78.9 &41.8 &39.9 &77.1 &44.6 &37.5\\
            Model C &75.6 &47.0 &40.2 &78.8 &46.2 &38.3 &67.3 &33.1 &26.6 &82.6 &52.1 &45.7 &78.9 &59.5 &47.4 &78.4 &40.4 &37.3 &76.9 &46.4 &39.2\\
            Model D &77.7 &54.9 &42.0 &81.2 &47.5 &40.0 &75.3 &45.3 &36.5 &82.0 &53.1 &44.4 &79.1 &62.2 &42.0 &80.6 &44.1 &40.9 &79.3 &51.2 &41.0\\
            \midrule
            Model E &77.3 &59.6 &42.4 &83.4 &52.0 &42.7 &77.4 &40.7 &31.1 &83.6 &53.5 &43.8 &80.2 &68.9 &42.9 &80.8 &43.8 &39.1 &80.5 &53.1 &40.3\\
            Model F &77.4 &51.4 &40.6 &83.9 &54.7 &46.6 &77.9 &47.9 &42.4 &83.4 &52.3 &44.5 &82.4 &65.2 &50.8 &83.3 &49.0 &45.0 &81.4 &53.4 &45.0\\
            Model G &80.4 &68.7 &45.4 &85.1 &56.4 &43.5 &78.7 &45.0 &35.3 &83.7 &49.4 &39.1 &84.9 &67.6 &50.6 &82.0 &46.7 &42.5 &82.5 &55.6 &42.7\\
            \midrule
            \textbf{GDRNet} &79.9 &66.8 &46.0 &84.7 &53.1 &45.3 &80.8 &45.3 &39.4 &84.0 &40.3 &35.9 &83.2 &63.4 &50.9 &82.9 &45.8 &43.5 &82.6 &52.5 &43.5\\
            \bottomrule
            \end{tabular}
    }}
    \label{tab:SDG}
    \vspace{-4mm}
\end{table*}


%
%
%